# LiteDiff: Lightweight Diffusion Model Adaptation for Data-Constrained Domain

*Ruchir Namjoshi, Nagasai Thadishetty, Vignesh Kumar, Hemanth Venkateshwara*
*Department of Computer Science, Georgia State University, Atlanta, USA*
*{rnamjoshi1, nthadishetty1, spandian1@student.gsu.edu}, hvenkateswara@gsu.edu*

## Abstract

*In recent years, diffusion models have demonstrated remarkable success in high-fidelity image synthesis. However, fine-tuning these models for specialized domains, such as medical imaging, remains challenging due to limited domain-specific data and the high computational cost of full model adaptation. In this paper, we introduce **LiteDiff (Lightweight Diffusion Model Adaptation)**, a novel fine-tuning approach that integrates lightweight adaptation layers into a frozen diffusion U-Net while enhancing training with a latent morphological autoencoder (for domain-specific latent consistency) and a pixel-level discriminator (for adversarial alignment). By freezing weights of the base model and optimizing only small residual adapter modules, LiteDiff significantly reduces the computational overhead and mitigates overfitting, even in minimal-data settings. Additionally, we conduct ablation studies to analyze the effects of selectively integrating adaptation layers in different U-Net blocks, revealing an optimal balance between efficiency and performance. Experiments on three chest X-ray datasets - **(1) Kaggle Chest X-Ray Pneumonia, (2) NIH Chest X-ray14** and **(3) VinBigData Chest X-ray**—demonstrate that LiteDiff achieves superior adaptation efficiency compared to naive full fine-tuning. Our framework provides a promising direction for transfer learning in diffusion models, facilitating their deployment in diverse low-data domains.*

## 1. Introduction

Diffusion-based generative models [6, 12, 20] have emerged as a powerful approach for generating high-fidelity images by progressively denoising an initial random noise distribution. These models effectively capture intricate details in real images, enabling state-of-the-art performance in text-to-image synthesis, as demonstrated by models like Stable Diffusion. The advent of latent diffusion models (LDMs) [17] has further enhanced their efficiency by operating in a compressed latent space, significantly reducing computational demands while maintaining high visual quality. However, when applying these broad-domain generative models to highly specialized fields such as medical imaging, direct fine-tuning presents several challenges.

**Motivation**: In many domain-specific applications—such as chest X-ray synthesis—the available training data is often limited to a few thousand samples, making direct adaptation of diffusion models both computationally prohibitive and prone to overfitting. Existing personalization techniques like DreamBooth [18] and Textual Inversion [3] primarily inject object-specific or concept-specific modifications but fail to reorient the generative model's distribution to conform to domain-specific constraints. In medical imaging, where morphology (structural shape and form) is a fundamental characteristic, naive adaptation strategies risk generating unrealistic or structurally inconsistent samples, thereby limiting their utility in downstream tasks.

**Key Idea**: To address these challenges, we propose **LiteDiff** — a framework that efficiently adapts pretrained diffusion models to specialized domains by integrating lightweight residual adaptation layers and enforcing domain-specific constraints. Instead of fine-tuning the entire UNet, LiteDiff incorporates small, residual adapter modules within selected blocks of the frozen UNet backbone, enabling effective adaptation with minimal additional parameters.

A key innovation in LiteDiff is the introduction of a Latent Morphology Autoencoder (LMA), designed to encode and enforce domain-specific structural consistency in the latent space. Unlike generic regularization approaches, LMA captures the essential morphological traits of the target domain—ensuring that the synthesized images preserve realistic anatomical structures. By introducing a latent consistency loss, LiteDiff aligns the generative process with the intrinsic morphological properties of real data, mitigating structural distortions commonly observed in naïve fine-tuning approaches. Additionally, we employ a pixel-level discriminator to further align generated images with the real domain distribution via adversarial learning.

**Contributions**: The main contributions of this work are as follows:
1. We propose LiteDiff, an efficient fine-tuning strategy for pre-trained diffusion models that integrates lightweight residual adapters, significantly reducing computational cost and mitigating overfitting risks.
2. We introduce the Latent Morphology Autoencoder (LMA), which learns domain-specific morphological representations and enforces a latent consistency constraint to ensure structural fidelity in generated images.
3. We conduct extensive experiments on chest X-ray datasets, including ablation studies on adapter placement and morphological constraints, demonstrating that Lite-Diff enables effective domain adaptation with minimal labeled data and computational resources.

## 2. Related Work

### 2.1. Diffusion Models for Image Generation

**Core Diffusion Approaches:**

Denoising Diffusion Probabilistic Models (DDPM) [6] introduced the core iterative denoising framework, followed by improvements such as improved sampling [12] and score-based SDEs [20]. Latent diffusion [17] further reduces the computational cost by performing the diffusion process in a learned latent space rather than pixel space. Such models are widely employed in text-to-image scenarios (*e.g.*, Stable Diffusion) due to their ability to handle high-resolution outputs efficiently.

**Diffusion in Specialized Domains:**

While diffusion has proven flexible, it typically requires large-scale training data. For specialized domains like medical imaging, certain works either train smaller diffusion models from scratch [21] or rely on unconditional approaches for data augmentation [25]. However, these rarely leverage the power of a large pre-trained text-conditioned model and often require domain-level or morphological constraints. Our method addresses this gap via minimal overhead adaptation and morphological-latent priors.

### 2.2. Transfer Learning and Personalization in Diffusion

**DreamBooth and Textual Inversion:**

Methods such as DreamBooth [18] and Textual Inversion [3] enable personalizing a diffusion model to represent specific subjects or concepts. They often fine-tune text embeddings or a small portion of the model so that new textual tokens correspond to user-provided images. However, these techniques do not necessarily shift the entire model distribution to a new domain. They are primarily targeted at adding or preserving concepts without major morphological or structural changes.

**LoRA and Adapter-Like Methods:**

Recent works on parameter-efficient fine-tuning include LoRA [8] and other low-rank updates to large language models. In the visual domain, some research explores injecting rank-limited transformations into the UNet's attention blocks. In general, these methods aim to drastically reduce trainable parameter overhead. Our approach is similar in spirit but focuses on a specific architecture of convolution-based residual adapters within each UNet block, combined with morphological-latent constraints for domain adaptation.

**Limited-Data Domain Transfer:**

Classical domain transfer in generative models often used smaller networks or strong regularization [23]. For diffusion, we see attempts that only update select submodules. Our morphological-latent penalty is an additional novelty to preserve structural patterns crucial in medical images.

### 2.3. Medical Image Synthesis and Shape Constraints

**GAN-Based Methods:**

Prior works on medical image synthesis commonly employed GANs for data augmentation or anonymization [2, 19]. They can produce realistic images but often fail at high resolution or require careful training to avoid mode collapse.

**Morphological Priors and Autoencoders:**

In the medical domain, morphological or anatomical priors can be essential, especially for tasks like organ segmentation or disease localization [1]. Including a domain autoencoder [29] or a cycle-consistency approach can enforce consistent geometry. Our latent morphology approach directly compares the real and generated images in the learned morphological embedding space, which helps avoid anatomically implausible results.

**Discriminator for Real-Fake Alignment:**

Adversarial approaches remain popular in medical imaging to ensure realism [27]. While diffusion models can produce plausible images, combining them with a domain-specific discriminator helps penalize domain-inconsistent artifacts. Our pixel-level disc. is relatively simple but effective when combined with shape-latent enforcement.

### 2.4. Adaptation Layers

Adaptation layers or residual adapters, are lightweight modules integrated into neural networks to adjust the pre-trained models to new tasks without extensive retraining [16]. By inserting these layers into specific parts of the network, models can adapt to new domains while retaining their original capabilities, thereby reducing the need for re-training large models from scratch.

## 2.5. Relevance to Our Work

Our method integrates the adaptation layers into a pre-trained model and employs adversarial training with a morphology penalty. This combination allows us to adapt the model to a new domain with limited data while maintaining the quality and diversity of generated images.

## 3. Methodology

In this work, we introduce **LiteDiff** — <u>Li</u>ghtweight <u>Diff</u>usion Model Adaptation for Data-Constrained Domains. Our objective is to adapt a large-scale, pre-trained diffusion model (Stable Diffusion) to a specialized domain (Chest X-ray dataset) using minimal training. The proposed framework consists of two phases of training:

- **Phase A:** Training a domain-specific latent morphology autoencoder. A vanilla autoencoder is trained to extract morphological domain features in its latent representation. This representation is used to guide the finetuning process during adaptation.
- **Phase B:** Reconfiguring a pre-trained diffusion model by inserting residual adapter layers in the UNet architecture and training these residual layers to generate images from the domain dataset.

**Notation**: We denote the training dataset images as $x \in \mathbb{R}^{C \times H \times W}$. The X-ray images can be gray scale ($C = 1$) or color ($C = 3$). $H$ and $W$ are image dimensions. The latent morphology autoencoder from Phase-A is denoted as LMA. The encoder in the LMA converts the input image to its latent morphology vector, with $z_{morph} \leftarrow \text{LMA.Enc}(x)$. The VAE component of the pre-trained Diffusion model is denoted as VAE. The UNet in the pre-trained Diffusion model denoted as $\text{UNet}(z_t, t, p)$, serves as the denoiser. Here, $t$ represents the time stamp, $z_t$ represents the noisy latent created from noisy image $x_t$, where $z_t \leftarrow \text{VAE.Enc}(x_t)$, and $p \leftarrow \text{CLIP.Enc}(y)$ is the CLIP embedding of an optional text prompt $y$.

### 3.1. Phase A: Latent Morphology Autoencoder

Medical images, like chest X-rays include vital structural details such as lung contours and rib patterns that must be accurately depicted in generated images. As our strategy is to freeze most of the diffusion model (including the UNet) and only update compact adapter modules, an auxiliary mechanism is needed to ensure that these domain-specific features are generated accurately. We pre-train a **Latent Morphology Autoencoder** on grayscale chest X-ray images to capture a dedicated morphological embedding. We use this embedding to guide the fine-tuning of the diffusion model.

**Setup**: Let $x$ denote a grayscale chest X-ray image [1]. A compact CNN-based encoder maps $x$ to a latent code, and a corresponding decoder reconstructs the image:

$$z_{morph} \leftarrow \text{LMA.Enc}(x), \quad \hat{x} \leftarrow \text{LMA.Dec}(z_{morph})$$

The LMA is trained to minimize the reconstruction error:

$$L_{LMA} = \|\hat{x} - x\| \quad (1)$$

**Freezing the Latent Morphology Autoencoder**: After the autoencoder converges (typically within 25–50 epochs on our limited dataset), we freeze its parameters. Subsequently, $\text{LMA.Enc}(.)$ provides a stable morphological embedding $z_{morph}$, which is used to penalize any structural discrepancies between real and generated images during the diffusion model adaptation phase.

### 3.2. Phase B: Diffusion Model Adaptation

In this phase, we adapt a pre-trained diffusion model by incorporating lightweight adapter layers into the frozen UNet. Figure 1 provides an overview of the process.

**Latent Diffusion Overview:** Following [17], a frozen VAE encodes a real image $x$ into a latent $z_0$. At a randomly selected time step $t$, Gaussian noise $\epsilon$ is added to form a noisy latent:

$$z_t \leftarrow \alpha_t z_0 + \sigma_t \epsilon \quad (2)$$

where $\alpha_t$ and $\sigma_t$ are derived from the diffusion schedule. The UNet then predicts the noise, $\hat{\epsilon} = \text{UNet}(z_t, t, p)$, where $p \leftarrow \text{CLIP.Enc}(y)$ is text encoding for $y$. The diffusion loss is defined as:

$$L_{recon} = \|\hat{\epsilon} - \epsilon\|_2^2 \quad (3)$$

During inference, the denoised latent is computed as $z_{pred} \leftarrow z_t - \hat{\epsilon}$ and decoded to produce the synthetic image.

$$x_{gen} \leftarrow \text{VAE.Dec}(z_{pred}) \quad (4)$$

#### 3.2.1. Adaptation Layers (Lightweight Residual Blocks)

**Motivation.** Fine-tuning the entire UNet is both computationally demanding and prone to overfitting when data is scarce. To address this, we insert small *residual adapters* — referred to as **Adaptation Layers** — at the end of each UNet block, keeping the original weights intact. Let $h^l$ denote the output of block $l$; the adapter then produces:

$$\tilde{h}^l \leftarrow h^l + \text{ReLU GN}(\text{Conv}_{1 \times 1}(h^l)) \quad (5)$$

---
[1] Although many chest X-ray datasets store images as 3-channel data for compatibility, the underlying information is grayscale; thus, the R/G/B channels are collapsed into one.

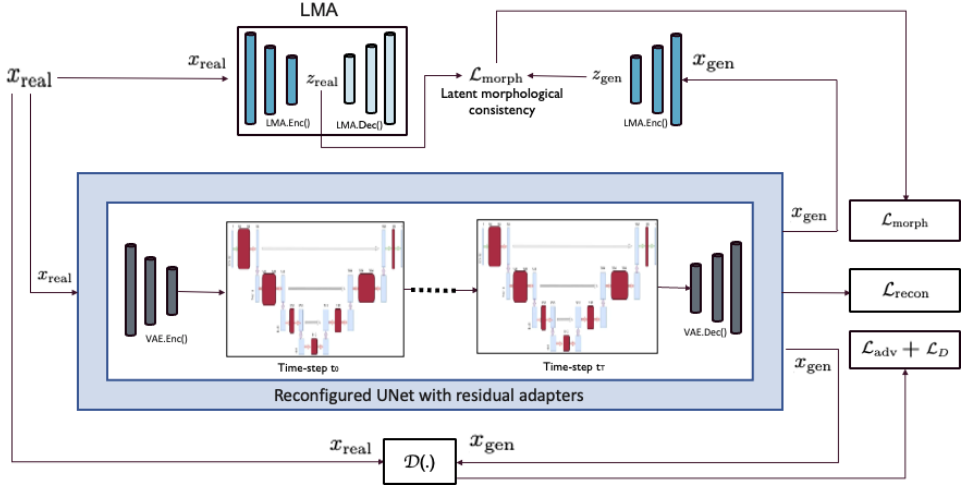
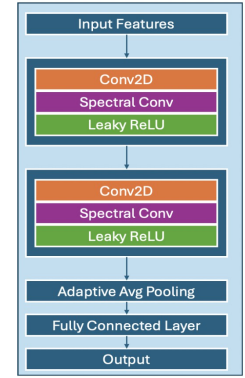

Figure 1. **Model Architecture.** Phase A: Train the Latent Morphological Autoencoder (LMA) and freeze its weights. Phase B: The frozen pre-trained UNet is reconfigured with adapter modules. During training, a mini-batch is encoded using the pretrained VAE, noise is added at a random time step $t$ and the UNet predicts $\hat{\epsilon}$. The denoised latent is decoded to yield $x_{gen}$, which is then used to compute both latent morphology by calculating the loss between latent space of $x_{real}$ and $x_{gen}$ using the LMA.Enc() and adversarial losses using the discriminator.

Figure 2. **Discriminator architecture.** Multiple convolutional layers with spectral normalization and ReLu Activation, followed by an adaptive pooling layer and a final convolutional layer that outputs a real/fake probability for $x_{real}$ and $x_{gen}$.

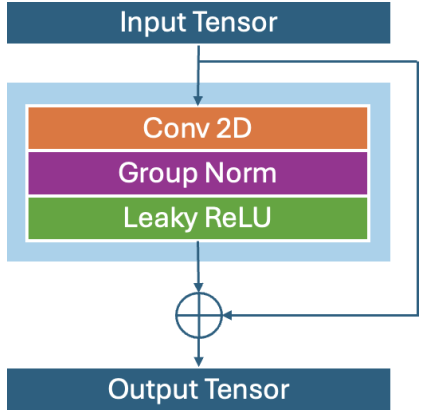

Figure 3. **Adapter layer structure.** A $1 \times 1$ convolution is followed by Group Normalization and ReLU activation, and the output is added residually to the original feature map. Only the parameters in these layers are trainable.

These adapters adjust the feature representations to align with the target domain without modifying the majority of the pre-trained model parameters. As shown in Figure 3, each adapter includes:

- A **Convolutional Layer** (with kernel size $1 \times 1$) that new feature representations.
- A **Group Normalization Layer** [26] that standardizes the features for domain alignment.
- A **Leaky ReLU Activation Function** [11] introducing non-linearity.

- A **Residual Connection** [4] that retains the original feature structure.

Only the parameters in these adapter modules (the $1 \times 1$ convolution, GroupNorm, and any ReLU affine parameters) are updated during training. The remaining parts of the UNet, including convolutional and attention modules, remain frozen.

**Implementation via Hooks.** As depicted in Figure 4, we attach the adapter modules using PyTorch forward hooks, which intercept the output $h^l$ of a block, process it through the adapter, and return the modified output $\tilde{h}^l$. This method enables us to selectively adapt specific blocks (for instance, omitting the mid-block or only adapting alternate blocks) without altering the underlying UNet code.

### 3.2.2. Pixel-Level Discriminator

After decoding $z_{pred}$ to obtain $x_{gen}$ = VAE.Dec($z_{pred}$), we employ a *pixel-level discriminator* $D(.)$ to differentiate between real and generated images. Designed as a multi-layer convolutional network with Spectral Normalization [10] (see Figure 2), the discriminator penalizes domain inconsistencies. The loss for the discriminator is given as follows.

$$L_D = \frac{1}{2} \left[ \text{BCE}\left(D(x_{real}), 1\right) + \text{BCE}\left(D(x_{gen}), 0\right) \right] \quad (6)$$

$$L_{adv} = \text{BCE}\left(D(x_{gen}), 1\right) \quad (7)$$

where BCE(.) denotes the binary cross-entropy loss. Minimizing $L_{adv}$ guides the adapters to produce features that ultimately generate realistic images, while $D(.)$ is updated using $L_D$.

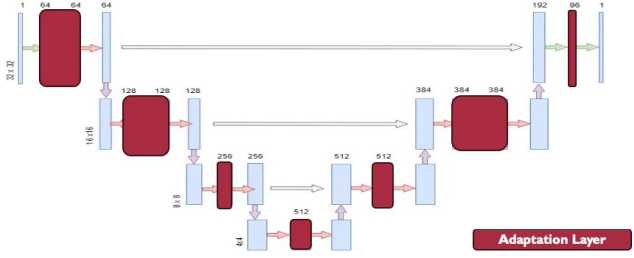

Figure 4. **Data Flow Through Unet Architecture.** Input data goes through the down blocks with sequential reduction in size, then flow through mid blocks and then through up blocks where they are concatened with output of corresponding down clocks to recreate the image dimensions. Adapter layers are inserted via hooks between layers of Unet Architecture.

### 3.2.3. Latent Morphology Autoencoder (LMA)

To further ensure that the generated images preserve the anatomical structure of chest X-rays, we enforce the latent morphology penalty using the Latent Morphology Autoencoder (LMA). The process is outlined below:

1. Obtain the morphological embeddings $z_{real} \leftarrow \text{LMA.Enc}(x_{real})$ and $z_{gen} \leftarrow \text{LMA.Enc}(x_{gen})$ using the LMA.
2. Apply an $L_2$ penalty to the difference.

$$L_{morph} \leftarrow \|z_{real} - z_{gen}\|_2^2 \quad (8)$$

Minimizing this loss encourages $x_{gen}$ the model to maintain key structural features present in the real images, preventing anatomical errors such as incorrect lung boundaries.

### 3.3. Overall Loss and Training Procedure

In Phase A, we train an auto-encoder (whose encoder part is used later for enforcing latent morphology consistency between real and generated images) and freeze its weights. For Phase B, we perform the steps for each mini-batch of chest X-rays, as represented in Algorithm 1, in a single training loop that updates both the discriminator and the adaptation modules.

**Total Objective Function:** The overall objective for adapting the diffusion model is determined as follows:

$$L_{gen} = L_{recon} + \lambda_{adv} L_{adv} + \lambda_{morph} L_{morph} \quad (9)$$

where $L_{recon}$ is the diffusion MSE loss, $L_{adv}$ promotes adversarial realism and $L_{morph}$ enforces morphological consistency. Hyper-parameters were set as $\lambda_{adv} = 0.1$ and $\lambda_{morph} = 0.001$, determined through rigorous trial and error.

---

**Algorithm 1** Training Process for Phase B of LiteDiff

**Require:** Training set of images $\{x\}$ and corresponding text description encoding using CLIP (optional) $\{p\}$, Pre-trained Diffusion VAE, Reconfigured Diffusion UNet with adapter residuals, Pre-trained morphology autoencoder LMA, pixel-level discriminator $D(.)$, and hyperparameters $\lambda_{adv}$, $\lambda_{morph}$

1: **for** each image $x_{real} \in \{x\}$ and text embedding $p$ **do**
2: $\quad z_0 \leftarrow \text{VAE.Enc}(x_{real})$
3: $\quad t \sim \text{Uni}[1, T], \epsilon \sim N(0, I)$
4: $\quad z_t \leftarrow \alpha_t z_0 + \sigma_t \epsilon$ Eq. 2
5: $\quad \hat{\epsilon} \leftarrow \text{UNet}(z_t, t, p)$
6: $\quad L_{recon} \leftarrow \|\hat{\epsilon} - \epsilon\|_2^2$ Eq. 3
7: $\quad x_{gen} \leftarrow \text{VAE.Dec}(z_t - \hat{\epsilon})$ Eq. 4
8: $\quad L_D \leftarrow \frac{1}{2} \text{BCE}(D(x_{real}), 1) + \text{BCE}(D(x_{gen}), 0)$
$\quad$ Update $D(.)$ using $\nabla L_D$
9: $\quad L_{adv} \leftarrow \text{BCE}(D(x_{gen}), 1)$ Eq. 7
10: $\quad L_{morph} \leftarrow \|z_{real} - z_{gen}\|_2^2$ Eq. 8
11: $\quad L_{gen} \leftarrow L_{recon} + \lambda_{adv} L_{adv} + \lambda_{morph} L_{morph}$
$\quad$ Update adapter modules in UNet using $\nabla L_{gen}$
12: **end for**

---

**Outcome:** By training only $1 \times 1$ adapter layers (approximately 3–4% of the UNet parameters) alongside a pixel-level discriminator, our method achieves efficient domain adaptation. This strategy not only accelerates the training process but also mitigates the risk of overfitting, making it particularly suitable for scenarios with limited training data.

### 3.4. Image Generation

At inference time, our goal is to generate images from scratch. The process begins by sampling a latent from a Standard Gaussian distribution, which we denote as $z_T$ (the noisiest latent). This latent is then iteratively refined through the reverse diffusion process using the adapted UNet and the diffusion scheduler. Specifically, the generation procedure is as follows:

1. **Initialization:** Sample an initial latent $z_T \sim N(0, I)$ of appropriate shape (typically matching the UNet's input dimensions).
2. **Text Conditioning:** Encode the desired text prompt $p$ using the text encoder, for e.g., CLIP(.) to obtain a prompt embedding.
3. **Reverse Diffusion:** Iteratively update the latent using the UNet with adapter modules. At each time step, the UNet predicts the noise component and the scheduler refines the latent.

$$z_{t-1} \leftarrow \text{Step}(z_t, \hat{\epsilon}(z_t, t, p)) \quad (10)$$

Where the process continues until reaching a final, denoised latent $z_0$.

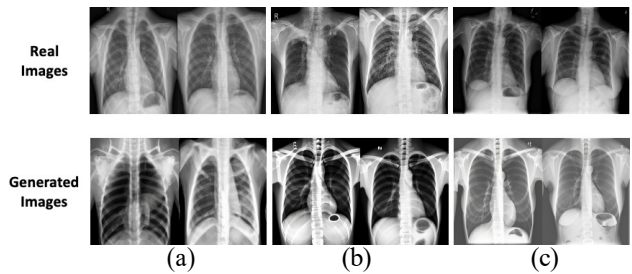

Figure 5. **Comparision with different datasets.** Our model is evaluated on three datasets: (a) Chest X-Ray Images (Pneumonia), (b) VinBigData Chest X-ray and (c) NIH Chest X-ray14. These datasets vary in chest X-ray types and dataset sizes.

| Method | FID ↓ | LPIPS ↓ |
|---|---|---|
| Chest X-Ray (Pneumonia) | 185.6 | 0.502 |
| Vin Big Dataset | 148.2 | 0.445 |
| Nih Dataset | **144.9** | **0.445** |

4. **Decoding:** Decode the final latent $z_0$ using the frozen VAE decoder to produce the generated image $x_{\text{gen}}$.

# 4. Experiments

We evaluate our **LiteDiff** — <u>Lite</u>weight <u>Diff</u>usion approach on three distinct chest X-ray datasets.
1. *Chest X-Ray Images (Pneumonia)* [9]: This is a public dataset sourced form Kaggle, consisting of 5,863 X-ray images categorized into two classes: Pneumonia and Normal.
2. *VinBigData Chest X-ray* [22]: This is a collection of 18K postero-anterior view X-ray scans, annotated with 6 labels for classifying and localization of common thoracic diseases.
3. *NIH Chest X-ray14 [24]:* This dataset comprises of 112,120 X-ray images sourced from 30,805 distinct subjects. Each image is annotated with up to 14 thorax disease labels.

We train our model on the above datasets and also compare to several baselines, report quantitative metrics (**FID** [5] and **LPIPS** [28]), analyze hooking ablations, and contrast training resource requirements versus prior medical-generation methods.

## 4.1. Setup and Results on different datasets

**Data Preprocessing:**

We resize all X-rays to $256 \times 256$ resolution. For training our shape auto-encoder, we convert them to single-channel grayscale (collapsing R/G/B if needed).

**Training on different datasets:**

As observed in Figure 5, we train on different types of datasets with varying number of images. The model adapts to capture specific patterns from each dataset type, such as the presence of a device installed in the chest.

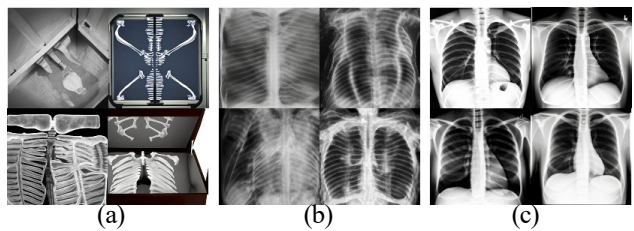

Figure 6. **Base Line Comparison.** This illustrates the outputs from: (a) pre-trained diffusion model, (b) our model without latent morphology consistency and (c) our final model.

## 4.2. Baselines Comparison

**Pre-Trained Stable Diffusion Vs After Domain Adaptation Method.**
- **Pre-trained Diffusion (No Adaptation):** Directly used stable diffusion (or an equivalent) off the shelf to generate "*a chest X-ray*"—effectively no domain shift.
- **No LMA:** We keep the adapter layers + adversarial disc but omit the latent morphology penalty. This tests the importance of morphological constraints.
- **LiteDiff:** Images generated using our model.

Table 1. **FID and LPIPS on chest X-ray data.** Lower is better. "No LMA" means no latent morphology penalty.

| Method | FID ↓ | LPIPS ↓ |
|---|---|---|
| Pre-trained diffusion | 341.8 | 0.86 |
| No LMA (only disc) | 215.2 | 0.69 |
| **LiteDiff (ours, all blocks)** | **148.2** | **0.445** |

## 4.3. Ablation Study

**Which Blocks to Adapt?**

We experiment with selectively hooking subsets of the UNet's *down*, *mid*, and *up* blocks:
- *All Blocks* (default)
- *Skip Down / Skip Mid / Skip Up*
- *Alternate Blocks* (hook indices 0,2,4,... )
- *Skip 2 Blocks* (hook indices 0,3,6,... )

Table 2 shows that adapting *all blocks* yields the best FID/LPIPS score. Skipping the entire segments (all *Down* blocks) increases FID. Meanwhile, hooking only the alternate layers shows moderate performance with reduced parameter overhead.

**How much data is enough ?**

We experiment with different amount of training data to understand minimal data needed to adapt to a new domain. We use *2000, 4000, 7500, 10000* and *15000* images from *Vin-BigData Chest X-ray* [22], measure the training time for 10 epochs and analyze the quality of images generated.

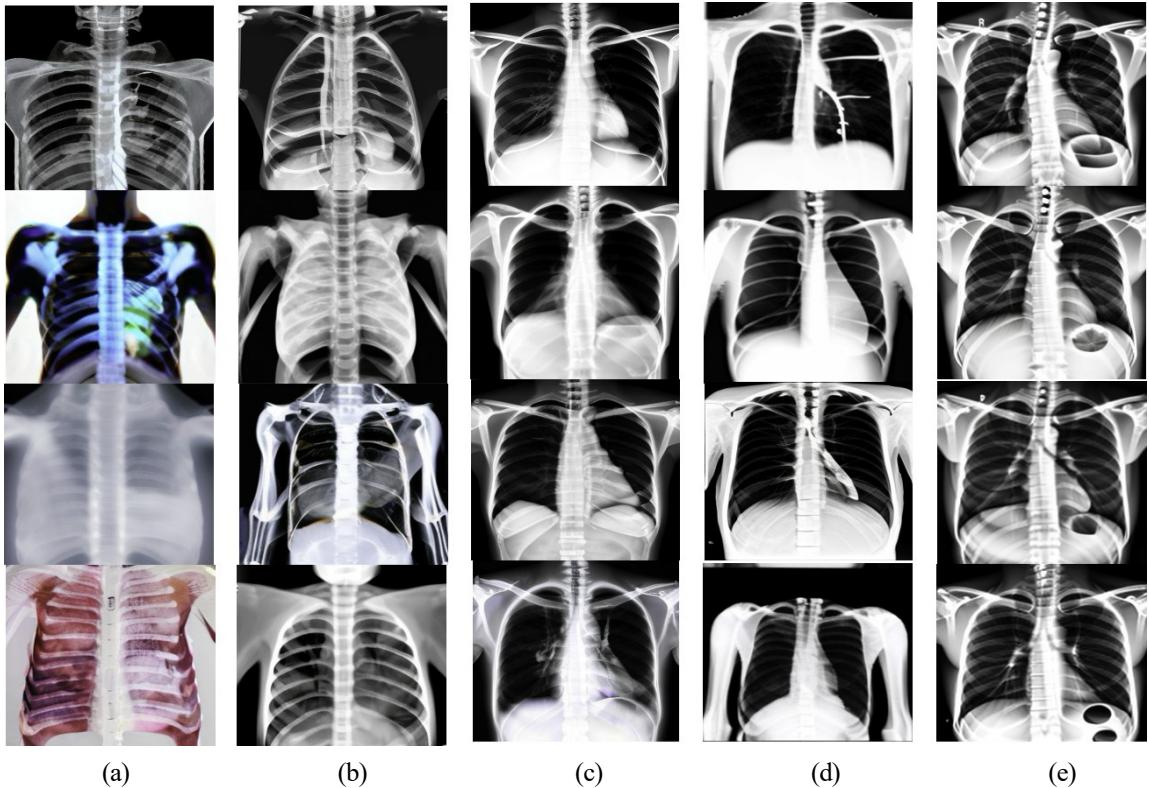

(a)          (b)          (c)          (d)          (e)

Figure 7. **Effect of training dataset size.** This visualization demonstrates the progressive improvement in image quality as training data volume increases across six configurations, from left to right: (a) 2,000 images, (b) 4,000 images, (c) 7,500 images, (d) 10,000 images, (e) 15,000 images and (f) 20,000 images.

Table 2. **Ablation on Hooking Patterns.** We vary which UNet blocks have adapters. "All Blocks" obtains best FID/LPIPS; skipping entire segments or hooking fewer blocks degrades performance slightly but reduces overhead.

| Hooking Pattern | FID ↓ | LPIPS ↓ |
|---|---|---|
| Skip Down | 267.211 | 0.648 |
| Skip Mid | 219.125 | 0.534 |
| Skip Up | 334.302 | 0.743 |
| Alternate Blocks | 186.576 | 0.543 |
| Skip 2 Blocks | 193.734 | 0.526 |
| Skip 1 only in Down | 187.982 | 0.500 |
| All Blocks | 148.2 | 0.445 |

| Data Size | FID ↓ | LPIPS ↓ | Training Time ↓ |
|---|---|---|---|
| 2000 Images | 262.1 | 0.71 | 40min |
| 4000 Images | 237.6 | 0.65 | 1hr 20 min |
| 7500 Images | 190.3 | 0.51 | 2hrs 30min |
| 10,000 Images | 163.9 | 0.473 | 3hrs 23min |
| 15,000 Images | 148.2 | 0.445 | 5hrs 04min |

Figure 7 shows an improvement in the quality of generated chest X-rays as the training dataset size increases. Notably, after training with 15,000 images, the model captures finer details, including the presence of a medical device in the chest X-ray, reflecting patterns learned from the training data.

**Quantitative Results.**

In Table 1, we compare different methods on our chest X-ray subsets. The *Pre-trained Diffusion (No Adaptation)* approach yields high FID (> 340), as it fails to generate anatomically consistent radiographs from a model trained on broad internet images. Removing the LMA degrade results. Our **LiteDiff (all blocks)** obtains a much lower FID (148.2) and lower LPIPS (0.445), indicating better distribution alignment and perceptual fidelity.

**Visual Analysis.**

Figure 6 provides the sample outputs. The pre-trained model alone often produces random artifacts or drastically incorrect lung geometry. Our latent morphology penalty en-sures robust morphological details, and the adversarial disc further refines texture consistency.

Table 3. **Model comparison with published methods.** "Trainable Params" is fraction of total parameters updated. Time durations are approximate and can vary by setup.

| Method | GPU | Trainable | Time |
| --- | --- | --- | --- |
| Stable Diff. [17] | 256×32GB | 100% | 2–3 wks |
| DreamBooth [18] | 1×24GB | 10–20% | 2–4 hrs |
| LoRA Fine-Tune [8] | 1×20GB | 5–10% | hrs–days |
| Okur *et al*. [15] | 1×11GB | 100% | 20–30 hrs |
| Frid-Adar *et al*. [2] | 1×24GB | 100% | multi-day |
| **LiteDiff (ours)** | **1×16GB** | **3–4%** | **4-5 hrs** |

### 4.4. Efficiency and Training Time

**Parameter Overhead.**

We only update ∼ 3–4% of the full UNet parameters via $1\times 1$ adapter modules, plus a small discriminator. In contrast, naive fine-tuning can involve 100% of the model.

**Training Duration.**

Since the majority of weights are frozen, training converges faster. On a single 16GB GPU, *LiteDiff* typically runs 10–12 epochs on ∼15k images in under 5 hours.

### 4.5. Comparison of Training Resources and Time

Table 3 summarizes resource usage for published works vs. our **LiteDiff**. Full diffusion training [17, 21] requires dozens or hundreds of GPUs for weeks. DreamBooth [18] or LoRA [8] need fewer resources but still needs to update large portions of the model. GAN approaches for chest X-ray [2, 15] often train from scratch for many hours on a single GPU. Our method remains lightweight (3–4% trainable) and converges in under 5 hours for a few thousand images.

**Summary.** LiteDiff requires far fewer resources and training data than full fine-tuning or other methods. This makes it practical for real medical labs with limited data and limited compute, yet seeking anatomically consistent synthetic X-rays.

## 5. Discussion

### 5.1. Why Small Adapters Work.

Adapting only $1 \times 1$ convolution-based adapters is reminiscent of adapter tuning in large language models [7]. Each block's feature is nudged to the new domain distribution without disturbing the overall diffusion process.

### 5.2. Skips & Partial Hooks.

Hooking every block yields the best results, but skipping certain blocks is still viable for reducing trainable parameters. This suggests a progressive adaptation approach: only adapt a subset of layers, critical for domain transformation.

### 5.3. Limitations.

Our latent morphology autoencoder relies on obtaining a stable and representative morphological embedding. In scenarios where the domain images are extremely scarce or exhibit high morphological variability, performance may degrade. In addition, while our method significantly reduces the trainable parameters, the overall FID score may not always be competitive with full fine-tuning or more specialized generative models, especially when the domain distribution is challenging. Furthermore, for radically different domains (e.g., color-coded functional scans vs. grayscale X-rays), a simple $1 \times 1$ convolution-based adapter may not be sufficient; more extensive adaptation or a combination of larger-scale modifications might be required to capture the necessary domain characteristics. Finally, the reliance on a single discriminator operating at the pixel level might limit the method's ability to capture higher-level structural or semantic nuances.

## 6. Conclusion

We have introduced **LiteDiff** — Lightweight Diffusion Model Adaptation for Data-Constrained Domains, a novel approach to adapt a pre-trained diffusion model to specialized domains with minimal data. By strategically inserting small, residual adapter layers into the frozen UNet, and enforcing latent morphology constraints and adversarial alignment, our framework achieves strong fidelity and preserves essential morphological features while updating only a minimal subset of the overall parameters. Our ablation studies further reveal that selective adaptation can reduce computational overhead with only a modest trade-off in performance.

Overall, **LiteDiff** enables domain adaptation of large diffusion models in resource-constrained settings and possibility of future extension to other medical or niche domains where full retraining is impractical.

## A. Appendix

To evaluate the effectiveness of **LiteDiff** in adapting to non-medical data, we conduct additional experiments by training it on two distinct datasets: (a) 10,000 Poké´mon images[14] and (b) Food-101[13]. The results are presented in Figure 9 and Figure 8 respectively. The generated images achieve an LPIPS score of 0.58 and an FID score of 92.5 for the Poké´mon dataset, while the Food-101 dataset yields an LPIPS score of 0.45 and an FID score of 102.638.

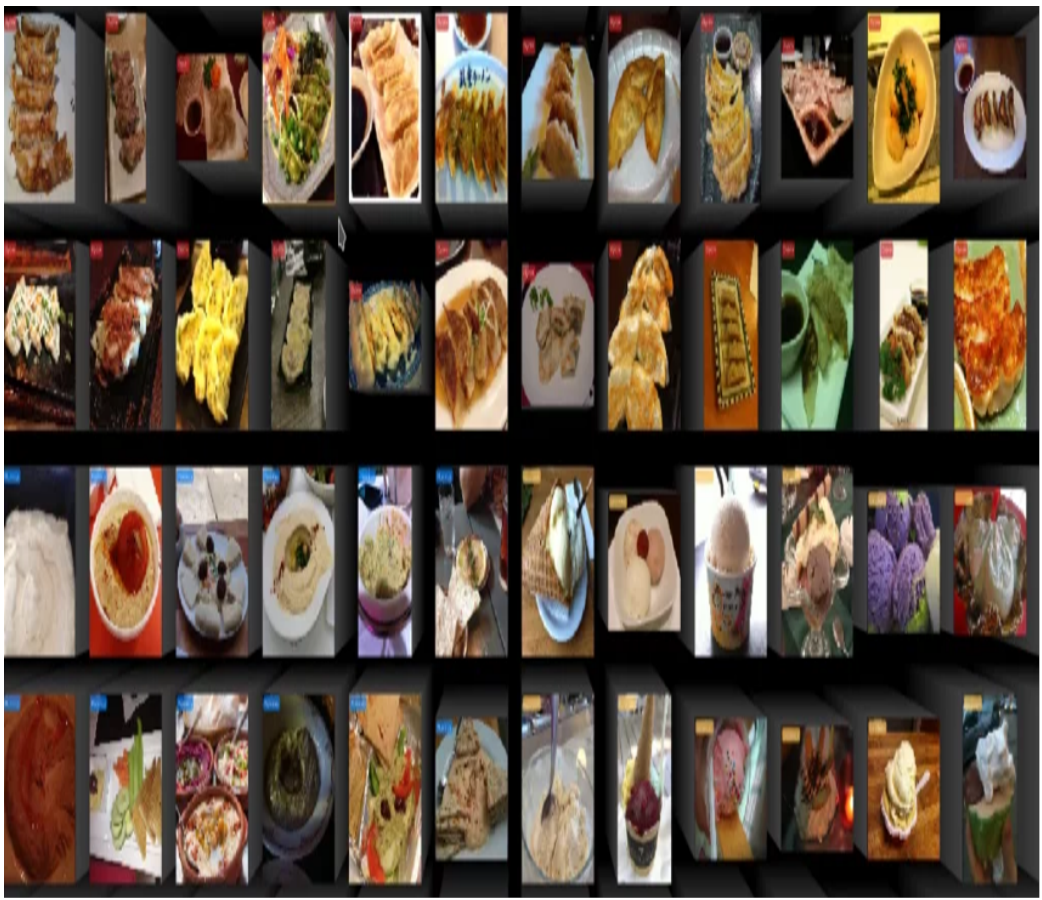

Figure 8. **Generated food images.**

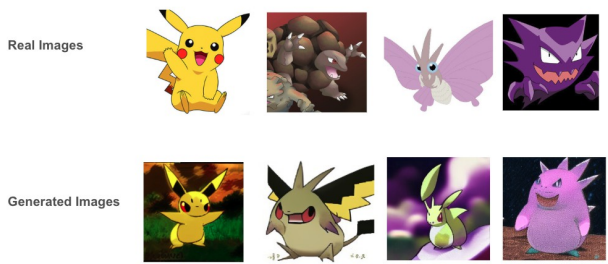

Figure 9. **Generated poké´mon images.**